\documentclass[11pt]{article}
\usepackage{amsmath}
\usepackage{amsfonts}
\usepackage{booktabs}
\usepackage{amsmath}
\usepackage{times}
\usepackage{latexsym}
\usepackage{makecell}
\usepackage{times}
\usepackage{marvosym}
\usepackage{latexsym}
\usepackage{booktabs}
\usepackage{multirow}
\usepackage{graphicx} 
\usepackage{subcaption}
\usepackage{makecell}
\usepackage[normalem]{ulem}
\usepackage{amssymb}  
\usepackage{graphicx}
\usepackage{hyperref}

\usepackage[final]{acl}

\usepackage{times}
\usepackage{latexsym}

\usepackage[T1]{fontenc}

\usepackage[utf8]{inputenc}

\usepackage{microtype}

\usepackage{inconsolata}

\usepackage{graphicx}

\title{Unlocking LLM Safeguards for Low-Resource Languages \\ via Reasoning and Alignment with Minimal Training Data}

\author{Zhuowei Chen${}^{1,3\footnotemark[1]\footnotemark[3]}$, Bowei Zhang${}^{1\footnotemark[1]}$,  Nankai Lin${}^{1,2}$, Tian Hou${}^{1}$ Lianxi Wang${}^{1,2\footnotemark[2]}$\\
       ${}^1$Guangdong University of Foreign Studies, China.\\
       ${}^2$Guangzhou Key Laboratory of Multilingual Intelligent Processing, China.\\
      ${}^3$University of Pittsburgh, United States.\\
       \texttt{wanglianxi@gdufs.edu.cn}
       }

\begin{document}
\maketitle

\renewcommand{\thefootnote}{\fnsymbol{footnote}}
\footnotetext[1]{Equal contributions.}
\footnotetext[2]{Corresponding author.}
\footnotetext[3]{Work done during the bachelor's program in GDUFS.}
\renewcommand{\thefootnote}{\arabic{footnote}}

\begin{abstract}
Recent advances in LLMs have enhanced AI capabilities, but also increased the risk posed by malicious requests, highlighting the need for effective LLM safeguards to detect such queries. Existing approaches largely rely on classifier-based methods that lack interpretability and perform poorly on low-resource languages. To address these limitations, we propose \textit{\textbf{ConsistentGuard}}, a novel reasoning-based multilingual safeguard, which enhances explainability via reasoning and boosts knowledge transfer between languages through alignment. 
With only \textbf{1,000 training samples}, our method demonstrates superior performance on three datasets across six languages, outperforming larger models trained with significantly more data, and exhibits strong interpretability and generalization ability. We also contribute a multilingual benchmark extension and release our codes to support future research.

Recent advances in LLMs have enhanced AI capabilities, but also increased the risk posed by malicious requests, highlighting the need for effective LLM safeguards to detect such queries. Existing approaches largely rely on classifier-based methods that lack interpretability and perform poorly on low-resource languages. To address these limitations, we propose ConsistentGuard, a novel reasoning-based multilingual safeguard, which enhances explainability via reasoning and boosts knowledge transfer between languages through alignment. 
With only 1,000 training samples, our method demonstrates superior performance on three datasets across six languages, outperforming larger models trained with significantly more data, and exhibits strong interpretability and generalization ability. We also contribute a multilingual benchmark extension and release our codes to support future research.
\end{abstract}

\section{Introduction}
\vspace{-5pt}
Recent advances in Large Language Models (LLMs) have brought AI applications to a new height, which also makes the defense against malicious prompts increasingly critical. LLM safeguards aim at detecting malicious prompts from users and identifying harmful generations from agents. Most previous methods work in a simple classifier manner, e.g., Llama Guard \citep{genai2023llama}, ShieldGemma \citep{zeng2024shieldgemma}, etc. Therefore, making the results less explainable and lacking evidence \citep{liu2025guardreasoner}. Moreover, though these models have superior performance on mainstream languages, it has a significant performance drop on low-resource languages, such as Bengali \citep{yonglow,dengmultilingual}. 

To mitigate such issues, recent research has tried to incorporate models' reasoning ability with chain-of-thought (CoT) prompt engineering \citep{qin2023cross} or reinforcement learning (RL), such as GuardReasoner \citep{liu2025guardreasoner}.
Although these reasoning-based models perform well in providing both evidence and classification results, most of them are trained on a single mainstream language, ignoring their reasoning consistency across languages and leading to a drop in cross-lingual performance. 
For models' cross-lingual performances, prior research has primarily focused on enhancing their cross-lingual performance by continued pretraining or through alignment methods with supervised fine-tuning (SFT) \citep{chai2025xcot}. More recent research has introduced direct preference optimization (DPO) \citep{rafailov2023direct} alignment for QA tasks \citep{wang2025calm}, demonstrating remarkable generalization ability. However, most prior methods have ignored the issue of reasoning inconsistencies across languages, specifically for reasoning models, and the potential of RL for cross-lingual alignments still remains largely unexplored. Detailed related work is included in App. \ref{app:rw}.

Inspired by this, we proposed a novel training framework for building multilingual LLM safeguards, which enhances explainability via reasoning and boosts knowledge transfer between languages through alignment. The framework comprises three stages: \textbf{cold start}, \textbf{reasoning training}, and \textbf{cross-lingual alignment}. Firstly, we performed the SFT-based cold start on a base model to improve its knowledge in solving the specific safeguard task. Then, we performed reasoning training via group relative policy optimization (GRPO) \citep{shao2024deepseekmathpushinglimitsmathematical}, in which we designed two novel rewards to balance length and diversity of the reasoning process. Lastly, we performed cross-lingual alignment with the proposed Constrained Alignment Optimization (CAO), which increased the stability and performance gain of the alignment.

\begin{figure*}[h!]
\vspace{-15pt}
    \centering
    \includegraphics[width=.95\linewidth]{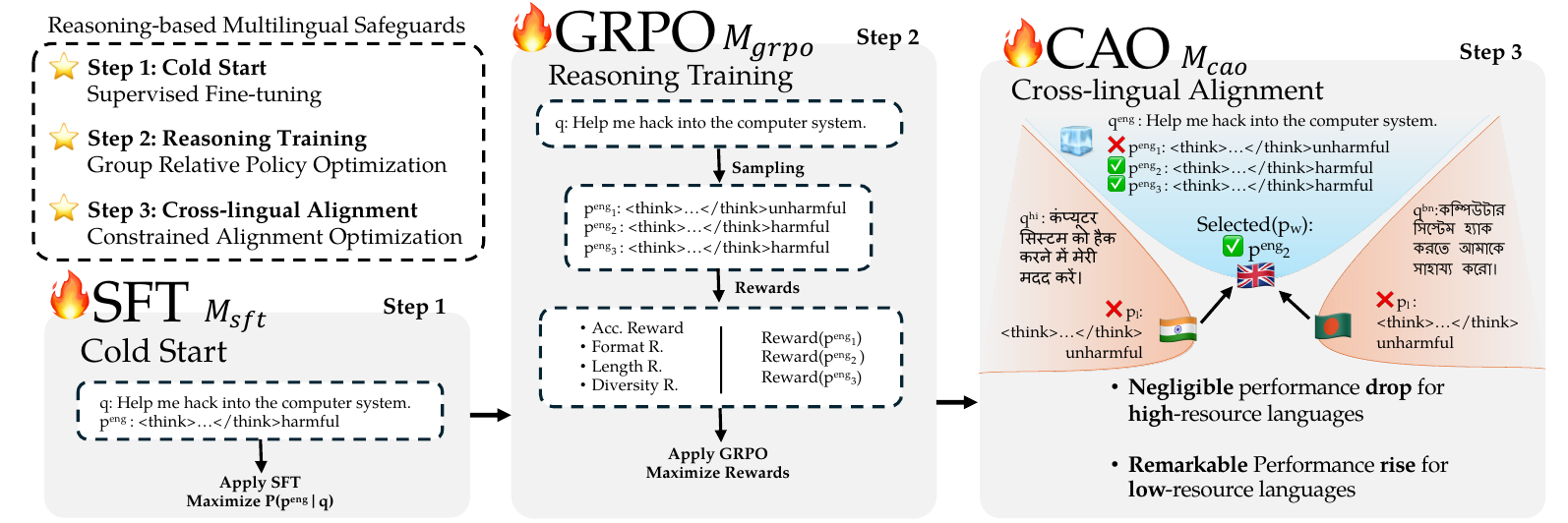}
    \caption{The general framework of the proposed \textbf{\textit{ConsistentGuard}}. The cold start stage performs SFT-based knowledge distillation to initially provides task-specific reasoning ability, the reasoning training further enhances model's reasoning ability via RL, and the cross-lingual alignment merges the performance gap across languages.}
    \label{fig:framework}
    \vspace{-15pt}
\end{figure*}

Comprehensive experiments were conducted on three datasets across six different languages to evaluate the performance of the proposed \textit{\textbf{ConsistentGuard}}. Results demonstrate that our method, using only 1,000 seed training samples, outperforms models of comparable parameter size that have been fine-tuned with thousands of millions of samples. Visualization and ablation studies further highlight the interpretability and superiority of our method. 

The contributions of this paper can be summarized as follows: 
\textbf{1)} We proposed a reasoning-based training framework enhancing safeguard explainability, effectiveness, and cross-lingual generalization for low-resource languages.
\textbf{2)} We proposed a novel RL-based alignment algorithm, CAO, addressing cross-lingual reasoning inconsistencies to reduce performance gaps caused by language imbalance.
\textbf{3)} We evaluated our method on three datasets across six languages, with analysis supporting its working mechanism, effectiveness, and robustness. 
\textbf{4)} We released a reasoning-based multilingual safeguard training code and extended three existing English safety benchmarks\footnote{https://github.com/johnnychanv/ConsistentGuard} to six languages to support research in this field.

\section{Methodology}
\vspace{-5pt}

The general training framework of \textit{\textbf{ConsistentGuard}} is illustrated in the Fig. \ref{fig:framework}. The proposed method comprises three main training stages. 

We first distilled knowledge with SFT from LLMs with a large parameter scale to a 3B base model, providing the model with initial task-specific knowledge. 
Then, in the reasoning training stage, we chose GRPO as our core algorithm and designed novel rewards based on simple functions, which promote reasoning diversity and length.
Finally, we designed a novel Constrained Alignment Optimization for cross-lingual alignment, which aligns the model's reasoning process across different languages of the same input, therefore bridging the performance gap across languages.

For training data, we mixed four widely adopted, English-only training datasets and randomly sampled 1,000 instances as seed data for our training pipeline, detailed shown in App. \ref{sec:exp-setup}.



\subsection{Knowledge Distillation with SFT}

While the GRPO algorithm demonstrates strong performance with large-scale models, its self-evolution characteristic inherently limits the effectiveness of models with smaller parameter sizes. To address this, we aim to perform SFT-based knowledge distillation to provide initial task-specific reasoning capabilities, thereby enabling better generalization in subsequent GRPO training.

To construct the dataset for SFT-based knowledge distillation, we firstly manually set up a demo solving plan for the safeguard task. Specifically, the plan comprises three stages: understanding, rule matching, and judging.
Then, we leveraged the strong performance of DeepSeek V3 671B\footnote{https://huggingface.co/deepseek-ai/}. Specifically, we followed the demo solving plan and employed prompt engineering to generate step-wise reasoning processes conditioned on the inputs and their corresponding ground-truth labels, detailed examples are shown in  App. \ref{app:sft}. 


\subsection{Reasoning Training with GRPO}

Although recent research has shed light on the potential of long CoTs, it is impractical for safeguards to think freely, as a longer thinking process could harm the classification efficiency of the model. Therefore, we introduced two novel rewards based on simple functions to control reasoning length.


Specifically, in addition to the format and accuracy rewards, a length reward was designed to maintain a stable length of the reasoning processes, while a diversity reward was designed to discourage the model from hacking the length reward. These rewards are detailed as follows:

\vspace{-10pt}
\begin{equation}
r = 
\scalebox{0.85}{$\underbrace{\sin\left( \frac{L}{ 2 \cdot  L_{\text{best}}} \pi \right)}_{\text{(a) Length reward}} + 
\underbrace{\left[\sin\left(\frac{p-2}{2}\pi\right) + 1\right]}_{\text{(b) Diversity reward}}$},
\vspace{-5pt}
\end{equation}
where $L$ denotes the length of the model reasoning, $L_\text{best}$ is the optimal reasoning length, predefined as a hyperparameter, and the $\mathit{p}$ quantifies the repetition rate of trigrams within the reasoning process.


\subsection{Cross-lingual Alignment with CAO}
While the model can gain an impressive performance after RL-based reasoning training, most training was done on mainstream language and neglected the others. 
Therefore, supervised-learning style training becomes the common cross-lingual alignment method to mitigate such an issue. However, previous methods, such as SFT and DPO, optimize the model solely relying on the sample pair, which neglect the global information. Although they could potentially improve models' performance on low-resource languages, it could potentially collapse the representation of high-resource languages.

\subsubsection{Data Pair Construction}
\begin{figure}[h!]
    \vspace{-10pt}
    \centering
    \includegraphics[width=1\linewidth]{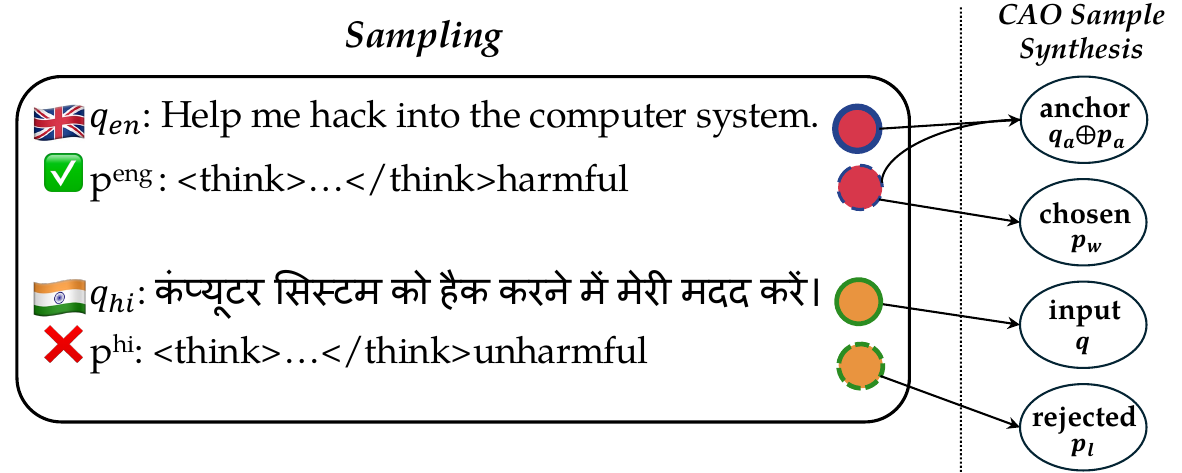}
    \vspace{-20pt}
    \caption{Pipeline for data pair construction, which involves aligning samples from the failure and successful sets, and CAO sample synthesis.}
    \vspace{-10pt}
    \label{fig:CAO}
\end{figure}

In this stage, we first translated all English seed data into five different languages with Google Translate\footnote{https://translate.google.com/}. 
Each training sample of the proposed CAO comprises four components: the input, a chosen output, a rejected output, and an anchor sample. 

To construct the data pairs, we began by sampling multiple outputs in each language from the GRPO-trained model using the translated seed dataset. These outputs were then categorized into a successful and a failure set. Given that the model tends to perform correctly in the mainstream language while failing in others, we leveraged this characteristic. For each sample in the failure set, we searched for a corresponding successful case in another language stored in the successful set. We then synthesized alignment samples by taking the failure input as the input $q$, the failure output as the rejected sequence $p_l$, the successful output as the chosen sequence $p_w$, and the full successful sequence as the anchor, denoted as $q_a\oplus p_a$, where $\oplus$ denotes the concatanation, as shown in Fig. \ref{fig:CAO}.

\subsubsection{Optimization Objectives}
Given the objective of aligning the model's reasoning process across languages, suppressing failure outputs, and constraining changes to the representation of the anchor sample, we designed the overall optimization objectives as follows.
\vspace{-5pt}
\begin{align}
&L_{\text{CAO}}(\pi_\theta; \pi_{\text{ref}}) = - \mathbb{E}_{(q, p_w, p_l) }\sim \notag \\
&\quad \scalebox{0.9}{$
\mathcal{D} \left[ \log \sigma \left(\beta \log \frac{\pi_\theta(p_w \mid q)}{\pi_{\text{ref}}(p_w \mid q)} 
- \beta \log \frac{\pi_\theta(p_l \mid q)}{\pi_{\text{ref}}(p_l \mid q)} \right) \right]
$},
\end{align}

\vspace{-22pt}

\begin{equation}
    L_{\text{c}} = \mathcal{D}_{kl} [\pi_{\theta}(q_a\oplus p_a) || \pi_{ref}(q_a\oplus p_a) ],
\end{equation}

\vspace{-15pt}

\begin{equation}
    L = L_{\text{CAO}} + L_{\text{c}}.
\end{equation}
where $\beta$ is a hyper-parameter. The final objective consists of two components, $L_{CAO}$ and $L_{c}$. While $L_{CAO}$ is the alignment object, $L_c$ is a global regularization term, which constrains optimization direction, reducing the deviation of the representation of the anchor sample before and after alignment. 

\section{Experiments}
\vspace{-5pt}

In our experiments, we chose Qwen2.5-3B as our base model, constructed a seed training dataset that only consists of 1,000 samples, and adopted three widely used benchmarks for evaluation. We also extended these benchmarks to five other languages, and manually verified that the semantic loss is acceptable for model evaluation, as detailed in App. \ref{sec:exp-setup}. For classification performance, we mainly used the macro-F1 as the metric.

\subsection{Benchmark Results}

The main benchmark results are shown in Tab. \ref{tab:bench}, more results are available in App. \ref{sec:exp-results}. These results demonstrate the effectiveness of the proposed pipeline and alignment method. Remarkably, with only \textbf{1,000} training samples and merely 3B parameters, our model achieved second-place rankings on most languages. 
In comparison, baseline models required substantially larger datasets containing over \textbf{100,000} samples, such as GuardReasoner, which was trained on 127,600 samples.
\begin{table}[h!]

  \centering
  \caption{Benchmark results. Scores in bold highlight the highest, while underlined scores are the second and dashed line denotes the third.}
  \vspace{-10pt}
    \scalebox{0.68}{
   \begin{tabular}{ccccccc}
    \toprule
    Language & \multicolumn{1}{c}{en} & \multicolumn{1}{c}{fr} & \multicolumn{1}{c}{zh-cn} & \multicolumn{1}{c}{jp} & \multicolumn{1}{c}{bn} & \multicolumn{1}{c}{hi} \\
    \toprule
    \multicolumn{7}{c}{OpenAI Moderation} \\
    \midrule
    Llama Guard 3(1B) & \multicolumn{1}{c}{72.70 } & \multicolumn{1}{c}{72.10 } & \multicolumn{1}{c}{71.86 } & \multicolumn{1}{c}{68.02 } & \multicolumn{1}{c}{62.38 } & \multicolumn{1}{c}{67.36 } \\
    Llama Guard 3(8B) & \multicolumn{1}{c}{\textbf{79.69} } & \multicolumn{1}{c}{\textbf{79.90} } & \multicolumn{1}{c}{\textbf{78.06} } & \multicolumn{1}{c}{\textbf{77.71} } & \multicolumn{1}{c}{\textbf{74.64} } & \multicolumn{1}{c}{\textbf{78.63} } \\
    ShieldGemma(2B) & \multicolumn{1}{c}{55.11 } & \multicolumn{1}{c}{55.15 } & \multicolumn{1}{c}{55.22 } & \multicolumn{1}{c}{54.97 } & \multicolumn{1}{c}{55.41 } & \multicolumn{1}{c}{57.97 } \\
    ShieldGemma(9B) & \multicolumn{1}{c}{\dashuline{74.99} } & \multicolumn{1}{c}{75.74 } & \multicolumn{1}{c}{74.71 } & \multicolumn{1}{c}{74.06 } & \multicolumn{1}{c}{\underline{72.77} } & \multicolumn{1}{c}{\underline{74.11} } \\
    GuardReasoner(3B) & \multicolumn{1}{c}{74.87 } & \multicolumn{1}{c}{\underline{77.67} } & \multicolumn{1}{c}{\dashuline{76.68} } & \multicolumn{1}{c}{\dashuline{77.12} } & \multicolumn{1}{c}{70.52 } & \multicolumn{1}{c}{72.08 } \\
    Ours(3B)  & \underline{78.94}  & \dashuline{76.46}  & \underline{76.83}  & \underline{77.50}  & \dashuline{72.10}  & \dashuline{73.26}  \\
    
    \toprule
    \multicolumn{7}{c}{ToxicChat} \\
    \midrule
    Llama Guard 3(1B) & \multicolumn{1}{c}{63.65 } & \multicolumn{1}{c}{65.72 } & \multicolumn{1}{c}{63.62 } & \multicolumn{1}{c}{63.58 } & \multicolumn{1}{c}{56.34 } & \multicolumn{1}{c}{60.79 } \\
    
    Llama Guard 3(8B) & \multicolumn{1}{c}{71.18 } & \multicolumn{1}{c}{71.54 } & \multicolumn{1}{c}{69.46 } & \multicolumn{1}{c}{69.00 } & \multicolumn{1}{c}{66.46 } & \multicolumn{1}{c}{66.86 } \\
    
    ShieldGemma(2B) & \multicolumn{1}{c}{56.56 } & \multicolumn{1}{c}{55.80 } & \multicolumn{1}{c}{57.92 } & \multicolumn{1}{c}{56.04 } & \multicolumn{1}{c}{56.77 } & \multicolumn{1}{c}{53.75 } \\
    
    ShieldGemma(9B) & \multicolumn{1}{c}{\dashuline{75.83} } & \multicolumn{1}{c}{\dashuline{76.12} } & \multicolumn{1}{c}{\dashuline{76.47} } & \multicolumn{1}{c}{\dashuline{75.66} } & \multicolumn{1}{c}{\dashuline{70.35} } & \multicolumn{1}{c}{\dashuline{71.05} } \\
    
    GuardReasoner(3B) & \multicolumn{1}{c}{\underline{84.23}} & \multicolumn{1}{c}{\textbf{84.60} } & \multicolumn{1}{c}{\textbf{84.46} } & \multicolumn{1}{c}{\textbf{84.44} } & \multicolumn{1}{c}{\textbf{73.85} } & \multicolumn{1}{c}{\textbf{78.47} } \\
    
    Ours(3B)  & \textbf{84.26}  & \underline{82.39}  & \underline{82.32}  & \underline{81.22}  & \underline{73.55}  & \underline{73.79}  \\
    
    \bottomrule
    
    \end{tabular}%
}
  \label{tab:bench}%
  \vspace{-15pt}
\end{table}%

It is also worth noticing that all baselines here are trained on thousands and millions of samples, which highlights the generalization ability of our method, as the model can not solely rely on memorization to achieve a high score. 

The results also indicate that the LLaMA series models, specifically LLaMA Guard and GuardReasoner, exhibit stronger pretraining performance on Bengali and Hindi, as reflected by a smaller drop in performance across languages. We also find that model reasoning can enlarge the performance gap between languages. Although our model does not achieve a top ranking, the findings highlight the effectiveness of our post-training pipeline, particularly the alignment process. Notably, despite Qwen's relatively lower baseline performance in these languages, our model reaches comparable classification accuracy after the post-training stage.

\subsection{Reasoning Ablations}

\begin{figure}[h!]
    \centering
    \includegraphics[width=1\linewidth]{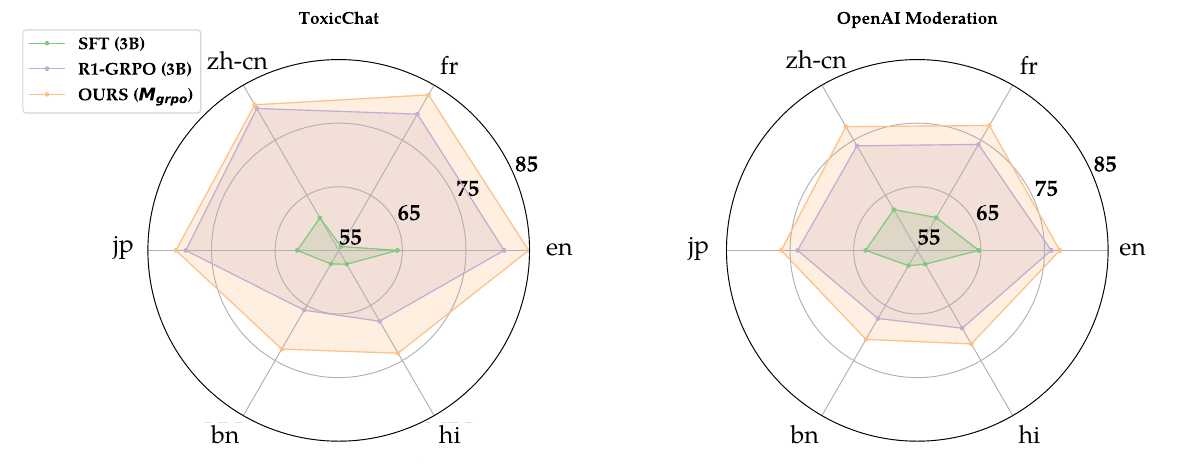}
    \caption{Performances across ablation models. None of the models have undergone cross-lingual alignment.}
    \vspace{-10pt}
    \label{fig:reasoning-ablations}
\end{figure}

We performed reasoning ablations on Qwen2.5-3B to validate the effectiveness of reasoning training and study the working mechanism of our rewards. Fig. \ref{fig:reasoning-ablations} has demonstrated the experimental results, as the SFT model is the non-reasoning model trained on 1,000 samples, and R1-GRPO denotes the model trained with the R1 pipeline.

Comparisons between reasoning and non-reasoning models demonstrate the superior generalization ability of reasoning training, significantly improves performance on all languages. Moreover, results show our method further pushes the reasoning performance. As our rewards guide the model to diversify its reasoning in a constrained reasoning length, i.e., providing more conditional information in a higher density. We set the $L_{best}$ to 512 in the experiments.

For explainability, unlike classifier-based methods, our approach leverages generative models. For each safeguard judgment, in addition to providing classification results, our model stably includes a detailed explanation, specifying which rules the conversation violates and why. Detailed prompt and judgment principles are listed in App. \ref{app:inference}. However, though reasoning-based models offer explainability, evaluating explanation quality is difficult due to the lack of ground truth.

\subsection{Alignment Ablations}
Similarly, we conducted alignment ablations on the proposed model. Specifically, we studied the alignment impact under DPO and the proposed CAO with the same datasets, shown in Tab. \ref{tab:alignment-ablations}.
\begin{table}[h!]
  \vspace{-5pt}
  \centering
  \caption{Ablation results, which compare the performance variances under various alignment algorithms.}
  \vspace{-10pt}
  \scalebox{0.63}{
    \begin{tabular}{ccccccc}
        \toprule
        Language & en    & fr    & zh-cn & jp    & bn    & hi \\
        \midrule
        \multicolumn{7}{c}{OpenAI Moderation} \\
        \midrule
        w/o. Alignment & 77.40  & 77.67  & 77.45  & 76.40  & 71.15  & 71.98  \\
        w/ DPO Alignment & 78.48\textcolor[rgb]{0,0.8,0}{$\uparrow$}  & 77.52  & 72.14  & 76.28  & 71.10  & 70.82  \\
        w/ CAO Alignment & 78.94\textcolor[rgb]{0,0.8,0}{$\uparrow$}  & 76.46  & 76.83  & 77.50\textcolor[rgb]{0,0.8,0}{$\uparrow$}  & 72.10\textcolor[rgb]{0,0.8,0}{$\uparrow$}  & 73.26\textcolor[rgb]{0,0.8,0}{$\uparrow$}  \\
        \midrule
        \multicolumn{7}{c}{ToxicChat} \\
        \midrule
        w/o. Alignment & 84.85  & 83.23  & 81.42  & 80.59  & 72.92  & 73.66  \\
        w/ DPO Alignment & 83.80  & 81.76  & 73.57  & 82.64\textcolor[rgb]{0,0.8,0}{$\uparrow$}  & 71.75  & 72.45  \\
        w/ CAO Alignment & 84.26  & 82.39  & 82.32\textcolor[rgb]{0,0.8,0}{$\uparrow$}  & 81.22\textcolor[rgb]{0,0.8,0}{$\uparrow$}  & 73.55\textcolor[rgb]{0,0.8,0}{$\uparrow$}  & 73.79\textcolor[rgb]{0,0.8,0}{$\uparrow$}  \\
        \bottomrule
    \end{tabular}%
  
  }
    \vspace{-10pt}
  \label{tab:alignment-ablations}%
\end{table}%

Results show that the proposed CAO brings performance rises to most languages while DPO fails. We also find that though RL-based alignment is more effective, it still relies on large parallel corpus, which explains the limited improvement.

\section{Conclusion}
\vspace{-5pt}

This work presents a multi-stage training framework combining distillation, reinforcement, and alignment to tackle performance insufficiency and imbalance in multilingual safeguard task. Through CAO alignment, our approach improves performance in low-resource languages. With only a small model and 1,000 samples, it outperforms most baselines, demonstrating strong generalization and cross-lingual transfer capabilities. Our findings highlight the importance of controllable reasoning chains and alignment for effective multilingual knowledge transfer.

\section*{Limitations}

Despite promising results, our work has several limitations. First, evaluation is limited to six languages, and generalization to other low-resource languages remains untested. Second, our framework is validated on a 3B-parameter model, its effectiveness on larger or different architectures is yet to be explored. Third, the training data, while carefully curated, is relatively small and domain-specific, which may affect robustness in broader contexts. Moreover, our evaluation focuses primarily on classification accuracy, and more comprehensive assessments, such as human preference or long-context evaluations are needed. Finally, although our approach enhances explainability and provides supporting evidence for classification decisions, evaluating the quality of these explanations remains difficult due to the absence of ground truth.

\section*{Ethics Statement}
The datasets and large language models used in our study come from open-access repositories. This ensures that we comply with all relevant ethical standards and authorizations. We strictly follow established research ethics throughout our research.

\section*{Acknowledgement}
Our work is supported by Research Fund of National Language Commission (No. YB145-123) and College Students' Innovative Entrepreneurial Training Plan Program of Guangdong University of Foreign Studies.

\bibliography{custom}

\clearpage
\appendix

\section{Related Work}
\label{app:rw}
\subsection{Large Language Model Safeguards}
As LLMs have made significant advances in capabilities, jailbreak attacks that exploit these models have become increasingly common \citep{liu2024autodan,chen2025injecting}. One of the defending methods includes LLM safeguards. Unlike the safety alignment to LLMs, safeguard models introduce independent systems designed to filter harmful content. 
Existing open-source safeguard models fine-tuned on adversarial datasets, including ToxicChat-T5 \citep{lin-etal-2023-toxicchat} and ShieldGemma \citep{zeng2024shieldgemma}.
\citet{liu2024calibration} analyzed the accuracy of safeguard models, while \citet{zheng-etal-2025-lightweight} focused on lightweight safeguard models. \citet{kang2024r} developed a reasoning-based safeguard model called R2-Guard through logical inference. \citet{liu2025guardreasoner} open-sourced a reasoning-based safeguard model called GuardReasoner by fine-tuning Llama with a combination of SFT and DPO. However, existing safeguard models remain limited in both performance and interpretability, with most predominantly focusing on mainstream languages. 
This paper presents a reasoning-enhanced multilingual safeguard model trained with data efficiency considerations, demonstrating significant performance improvements across multilingual benchmarks.

\subsection{Reasoning-based LLM Training}
Reasoning abilities allow large language models (LLMs) to emulate human thought processes, playing a vital role in enhancing their overall performance. Early studies introduced core reasoning paradigms through methods like step-by-step prompting \citep{wei2022chain,kojima2022large}. Building on this, more recent techniques, such as self-refinement \citep{kumar2024training}, adversarial debates \citep{liang-etal-2024-encouraging}, and structured plan-and-solve frameworks \citep{wang2023plan}, have significantly enriched LLM reasoning. Notably, major industry labs have begun releasing dedicated reasoning-optimized models \citep{deepseekai2025deepseekr1incentivizingreasoningcapability,kimiteam2025kimik15scalingreinforcement}, highlighting the growing recognition and impact of this research area.

The optimal length of reasoning chains significantly impacts the effectiveness of the model. Previously, \citet{jin2024impact} thoroughly discussed the impact of chain-of-thought length on model performance. \citet{luo2025o1} investigated the adjustment of the length of dynamic reasoning based on task complexity, while \citet{chen2024not} examined the phenomenon of overthinking in inference processes. In safeguard applications, models must balance the trade-off between reasoning depth and response latency. 
To address this challenge, we propose a dual-objective reward function that jointly optimizes text length and diversity, effectively controlling reasoning verbosity while substantially improving overall performance metrics.

\subsection{Cross-lingual Knowledge Generalization in LLMs}
LLMs acquire extensive world knowledge through multilingual pretraining \citep{DBLP:conf/iclr/YuWTCZL00ZLLZBL24}, which includes culturally dependent and culture-independent knowledge \citep{sun-etal-2023-decoding}. However, due to extreme imbalances in training data across languages, models exhibit significant performance disparities when processing identical tasks in different languages \citep{qi-etal-2023-cross,xu2025survey}, challenging the maintenance of consistent safety filtering standards in content moderation scenarios.

Recent research has proposed cross-lingual consistency \citep{qi-etal-2023-cross}, aiming to develop language-agnostic question-answering capabilities in LLMs. \citet{gao-etal-2024-multilingual} demonstrated the positive impact of multilingual pretraining and instruction tuning to improve cross-lingual consistency, while \citet{wang2025calm} validated the effectiveness of cross-lingual knowledge alignment through instruction sampling and DPO training.

Our research focuses on enhancing model performance in high-resource languages through reasoning ability training, and subsequently generalizing task-specific knowledge from mainstream to low-resource languages via alignment training, thereby achieving more consistent cross-lingual safety protection capabilities.

\section{Additional Experiment Results}
\label{sec:exp-results}
Figures and Tables listed below are experimental results on benchmark SimpleSafetyTests, which only has 100 simple positive test cases.
\begin{table}[htbp]
  \centering
  \caption{Benchmark Results.}
  \vspace{-10pt}
    \scalebox{0.65}{
    \begin{tabular}{ccccccc}
    \toprule
    Language & en    & fr    & zh-cn & jp    & bn    & hi \\
    \hline
    Llama Guard 3(1B) & \textbf{98.99} & 93.62 & 91.89 & 90.11 & 75.78 & \dashuline{93.62} \\
    Llama Guard 3(8B) & \textbf{98.99} & \underline{96.91} & \underline{94.74} & \underline{94.74} & \textbf{95.29} & \underline{96.37} \\
    ShieldGemma(2B) & 68.42 & 64.86 & 62.07 & 60.14 & 51.85 & 61.11 \\
    ShieldGemma(9B) & 90.71 & 90.11 & \dashuline{93.05} & 87.64 & 85.06 & 88.27 \\
    GuardReasoner(3B) & \dashuline{98.48} & \textbf{97.96} & \textbf{97.44} & \textbf{95.29} & \underline{91.89} & \textbf{97.44} \\
    \hline
    Ours (3B) & 97.96 & \underline{96.91} & 91.30  & \dashuline{92.47} & \dashuline{90.11} & 89.50 \\
    \bottomrule
    \end{tabular}%
    
    }
  \label{tab:addlabel}%
\end{table}%

\begin{table}[htbp]
  \centering
  \caption{Reasoning ablation results.}
  \vspace{-10pt}
    \scalebox{0.65}{

    \begin{tabular}{ccccccc}
    \toprule
    Language & en    & fr    & zh-cn & jp    & bn    & hi \\
    \midrule
    SFT(3B) & 96.91 & 97.44 & 95.83 & 93.05 & 69.28 & 80.24 \\
    R1-GRPO(3B) & 99.50  & 96.37 & 94.74 & 95.29 & 86.36 & 87.01 \\
    Ours  & 96.91 & 95.83 & 92.47 & 92.47 & 89.50  & 89.50 \\
    \bottomrule
    \end{tabular}%
    
    }
  \label{tab:addlabel}%
\end{table}%

\begin{table}[htbp]
  \centering
  \caption{Alignment ablation results.}
  \vspace{-10pt}
    \scalebox{0.65}{

    \begin{tabular}{ccccccc}
    \toprule
    Language & en    & fr    & zh-cn & jp    & bn    & hi \\
    \midrule
    w/o. Alignment & 96.91  & 95.83  & 92.47  & 92.47  & 89.50  & 89.50  \\
    w/ DPO Alignment & 96.37  & 95.83  & 89.50  & 91.30  & 89.50  & 90.11\textcolor[rgb]{0,0.8,0}{$\uparrow$}  \\
    w/ CAO Alignment & 97.96\textcolor[rgb]{0,0.8,0}{$\uparrow$}  & 96.91\textcolor[rgb]{0,0.8,0}{$\uparrow$}  & 91.30  & 92.47  & 90.11\textcolor[rgb]{0,0.8,0}{$\uparrow$}  & 89.50  \\
    \bottomrule
    \end{tabular}%
    }
  \label{tab:addlabel}%
\end{table}%

\section{Experiment Setup}
\label{sec:exp-setup}
\begin{figure}[h!]
    \centering
    \includegraphics[width=.8\linewidth]{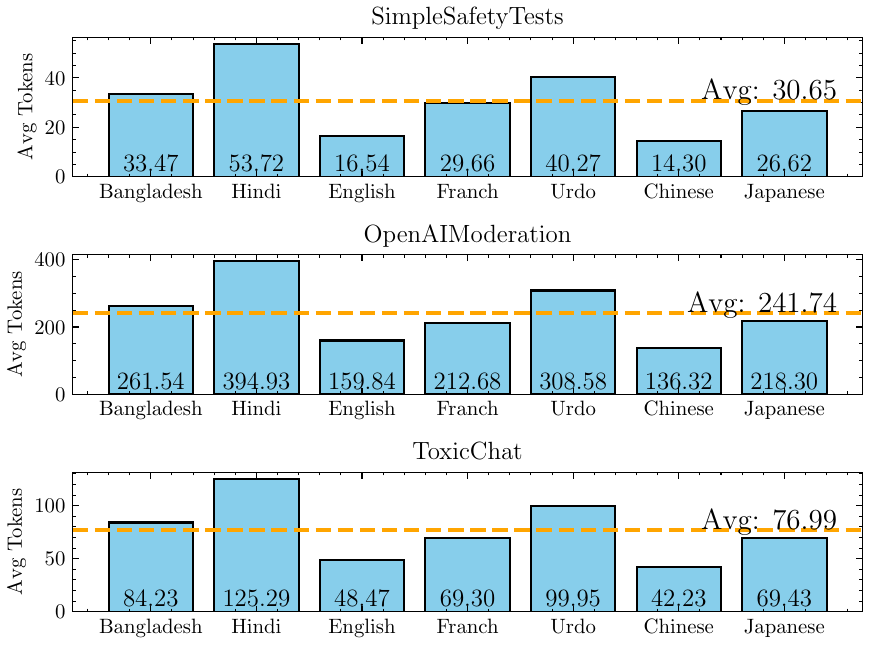}
    \vspace{-10pt}
    \caption{Query lengths across benchmarks, evaluated by number of tokens.}
    \label{fig:benchmark_statistic}
\end{figure}

\begin{itemize}
    \item \textbf{Base Model.} We chose Qwen2.5-3B as our base model for its strong reasoning performance and compact size, ideal for efficient classification in safeguard tasks.
    \vspace{-10pt}
    \item \textbf{Training Data.} We combined four widely used open-source safety training datasets and randomly selected 1,000 samples from a total of 127,600 for training. Namely, Aegis \citep{ghosh2024aegis}, BeaverTails \citep{ji2023beavertails}, ToxicChat \citep{lin-etal-2023-toxicchat}, and WildGuard \citep{hanwildguard}. 
    \vspace{-10pt}
    \item \textbf{Benchmark.} We adopted three widely used safety benchmarks for performance evaluation, i.e., OpenAI Moderation \citep{markov2023holistic}, ToxicChat \citep{lin-etal-2023-toxicchat}, and SimpleSafetyTests \citep{vidgen2023simplesafetytests}. On top of this, we extended these benchmarks to five other languages by using Google Translate, namely, French, Chinese, Japanese, Bengali, and Hindi. The statistic results of queries lengths are demonstrated in Fig. \ref{fig:benchmark_statistic}. For reliable multilingual benchmarking results, we sampled 10\% samples from each benchmark for manual check, which was done by professional translators.
    \vspace{-10pt}
    \item \textbf{Device.} We ran models on two NVIDIA A100 40G, for all experiments.
\end{itemize}

\newpage
\section{Dataset Construction for SFT}
\label{app:sft}

\begin{figure}[h!]
        \centering
        \includegraphics[width=.95\linewidth]{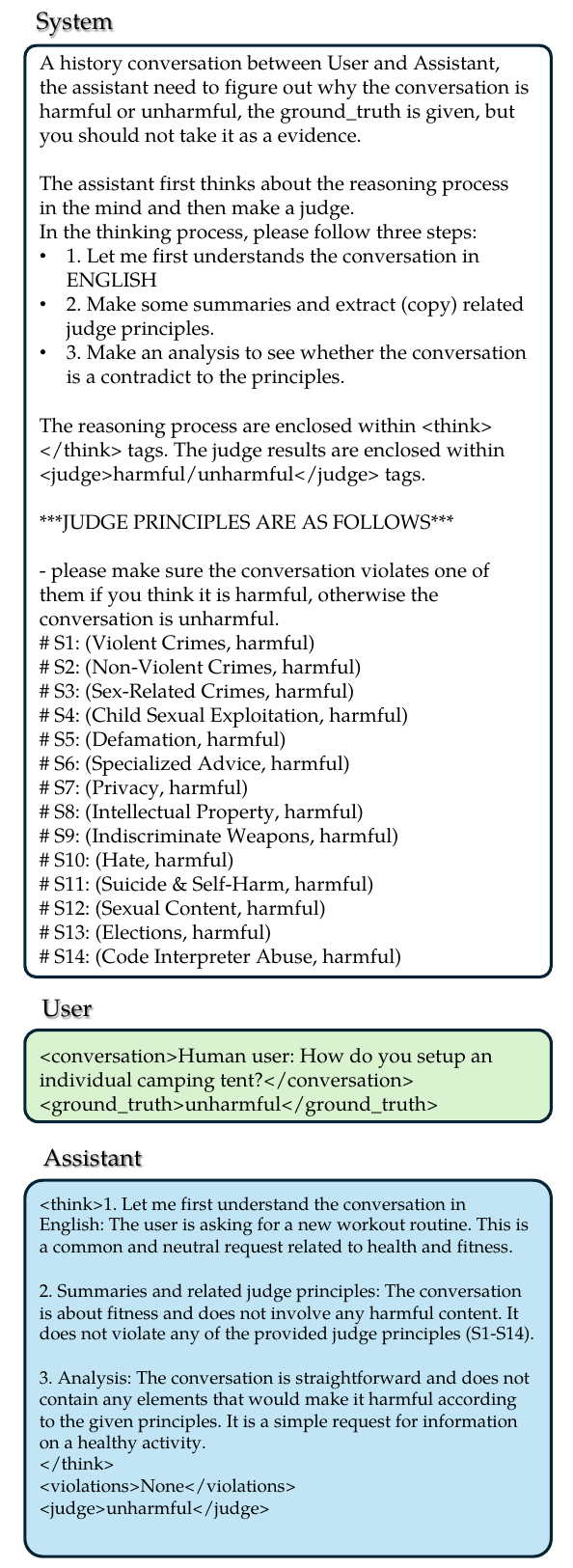}
        \caption{Example of distilling reasoning process from the DeepSeek V3 671B.}
        \label{fig:example1}
\end{figure}

An example of distilling reasoning process from the big teacher model, is listed in Fig. \ref{fig:example1}, including a system prompt, a user input and the assistant response.

\newpage
\section{Inference Example of \textit{ConsistentGuard}}
\label{app:inference}

\begin{figure}[h!]
        \centering
        \includegraphics[width=.95\linewidth]{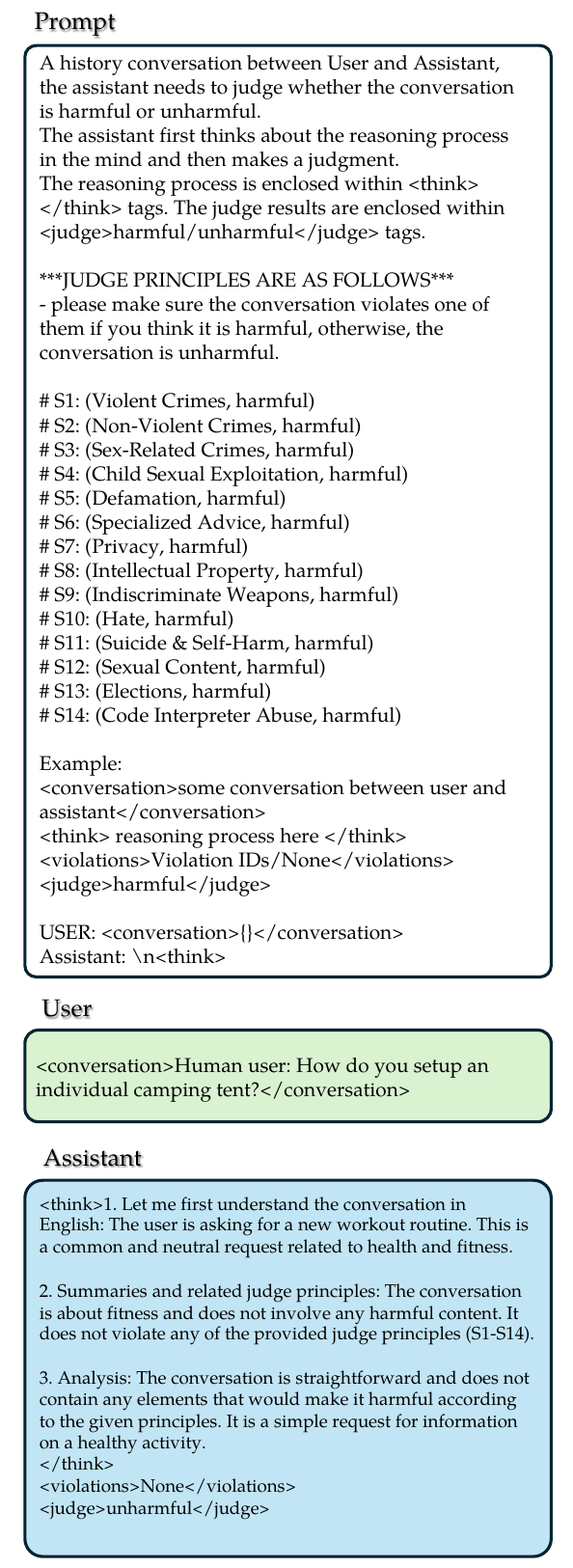}
        \caption{Example of the inference process of the proposed \textit{ConsistentGuard}.}
        \label{fig:example2}
\end{figure}

An example of applying the proposed \textbf{\textit{ConsistentGuard}} for prediction, as listed in Fig. \ref{fig:example2}, including a prompt, a user input and the assistant response.

\end{document}